\documentclass[fleqn,10.5pt]{qb}
\usepackage{xcolor}
\usepackage{framed}
\definecolor{shadecolor}{gray}{.85}

\usepackage{amsmath,amssymb,amsfonts}
\usepackage{algorithmic}
\usepackage{graphicx}
\usepackage{textcomp}
\usepackage{comment}
\usepackage{footnote}
\usepackage{threeparttable}
\usepackage{romannum}
\usepackage{longtable}
\usepackage{lscape}
\usepackage{soul}
\usepackage{array, makecell}
\usepackage{url}
\usepackage{tikz}
\usepackage{pgfplots}
\pgfplotsset{compat=newest}
\usepackage{pgfplotstable}
\usepackage{pgf-pie}
\usepackage{longtable}
\usepackage[colorlinks=true,linkcolor=blue,urlcolor=blue,citecolor=blue]{hyperref}



\title{A Survey of Machine Learning Techniques for Detecting and Diagnosing COVID-19 from Imaging}

\author[1]{Aishwarza Panday}
\author[2,*]{Muhammad Ashad Kabir}
\author[3]{Nihad Karim Chowdhury}
\affil[1]{Department of Computer Science \& Engineering, Stamford University, Dhaka 1217, Bangladesh}
\affil[2]{School of Computing and Mathematics, Charles Sturt University, NSW 2795, Australia}
\affil[3]{Department of Computer Science \& Engineering, University of Chittagong, Chittagong 4349, Bangladesh}
\affil[*]{corresponding author: akabir@csu.edu.au}

\begin{document}

\flushbottom
\maketitle
\textbf{ABSTRACT}

\noindent\textit{Background:} 
Due to the limited availability and high cost of the reverse transcription-polymerase chain reaction (RT-PCR) test, many studies have proposed machine learning techniques for detecting COVID-19 from medical imaging.
The purpose of this study is to systematically review, assess and synthesize research articles that have used different machine learning techniques to detect and diagnose COVID-19 from chest X-ray and CT scan images.

\noindent\textit{Methods:} A structured literature search was conducted in the relevant bibliographic databases to ensure that the survey solely centered on reproducible and high-quality research. We selected papers based on our inclusion criteria.

\noindent\textit{Results:} In this survey, we reviewed $98$ articles that fulfilled our inclusion criteria.
We have surveyed a complete pipeline of chest imaging analysis techniques related to COVID-19, including data collection, pre-processing, feature extraction, classification, and visualization.
We have considered CT scans and X-rays as both are widely used to describe the latest developments in medical imaging to detect COVID-19.

\noindent\textit{Conclusions:} 
This survey provides researchers with valuable insights into different machine learning techniques and their performance in the detection and diagnosis of COVID-19 from chest imaging.
At the end, the challenges and limitations in detecting COVID-19 using machine learning techniques and the future direction of research are discussed.

\vspace{1em}

\begin{shaded*}
    \noindent\textbf{Author summary:} This study provides researchers with valuable insights into different machine learning techniques and their performance to establish an automatic diagnosis system for COVID-19 using X-rays and CT-scans. 
\end{shaded*}

\section{Introduction}
\label{sec:introduction}

\begin{figure}[!t]
\centering
\resizebox{11cm}{!}{

\tikzset{every picture/.style={line width=0.75pt}} 

\begin{tikzpicture}[x=0.75pt,y=0.75pt,yscale=-1,xscale=1]

\draw    (152.5,68) -- (310.5,68) ;
\draw    (153,67) -- (153.45,84) ;
\draw [shift={(153.5,86)}, rotate = 268.49] [color={rgb, 255:red, 0; green, 0; blue, 0}  ][line width=0.75]    (10.93,-3.29) .. controls (6.95,-1.4) and (3.31,-0.3) .. (0,0) .. controls (3.31,0.3) and (6.95,1.4) .. (10.93,3.29)   ;
\draw  [fill={rgb, 255:red, 255; green, 255; blue, 255 }  ,fill opacity=1 ] (161.5,13) -- (298.5,13) -- (298.5,49) -- (161.5,49) -- cycle ;
\draw  [fill={rgb, 255:red, 255; green, 255; blue, 255 }  ,fill opacity=1 ] (85.5,88) -- (222.5,88) -- (222.5,124) -- (85.5,124) -- cycle ;
\draw    (311,67) -- (311.45,84) ;
\draw [shift={(311.5,86)}, rotate = 268.49] [color={rgb, 255:red, 0; green, 0; blue, 0 }  ][line width=0.75]    (10.93,-3.29) .. controls (6.95,-1.4) and (3.31,-0.3) .. (0,0) .. controls (3.31,0.3) and (6.95,1.4) .. (10.93,3.29)   ;
\draw  [fill={rgb, 255:red, 255; green, 255; blue, 255 }  ,fill opacity=1 ] (243.5,88) -- (380.5,88) -- (380.5,124) -- (243.5,124) -- cycle ;
\draw    (72.5,142) -- (230.5,142) ;
\draw    (73,142) -- (73.45,159) ;
\draw [shift={(73.5,161)}, rotate = 268.49] [color={rgb, 255:red, 0; green, 0; blue, 0 }  ][line width=0.75]    (10.93,-3.29) .. controls (6.95,-1.4) and (3.31,-0.3) .. (0,0) .. controls (3.31,0.3) and (6.95,1.4) .. (10.93,3.29)   ;
\draw    (230,141) -- (230.45,158) ;
\draw [shift={(230.5,160)}, rotate = 268.49] [color={rgb, 255:red, 0; green, 0; blue, 0 }  ][line width=0.75]    (10.93,-3.29) .. controls (6.95,-1.4) and (3.31,-0.3) .. (0,0) .. controls (3.31,0.3) and (6.95,1.4) .. (10.93,3.29)   ;
\draw  [fill={rgb, 255:red, 255; green, 255; blue, 255 }  ,fill opacity=1 ] (3.5,163) -- (140.5,163) -- (140.5,199) -- (3.5,199) -- cycle ;
\draw  [fill={rgb, 255:red, 255; green, 255; blue, 255 }  ,fill opacity=1 ] (162.5,163) -- (299.5,163) -- (299.5,199) -- (162.5,199) -- cycle ;
\draw    (228.5,197.75) -- (228.5,217) ;
\draw    (225.5,48.75) -- (225.5,68) ;
\draw    (144.5,123.75) -- (144.5,143) ;
\draw    (152.5,218) -- (310.5,218) ;
\draw    (152,217) -- (152.45,234) ;
\draw [shift={(152.5,236)}, rotate = 268.49] [color={rgb, 255:red, 0; green, 0; blue, 0 }  ][line width=0.75]    (10.93,-3.29) .. controls (6.95,-1.4) and (3.31,-0.3) .. (0,0) .. controls (3.31,0.3) and (6.95,1.4) .. (10.93,3.29)   ;
\draw    (311,217) -- (311.45,234) ;
\draw [shift={(311.5,236)}, rotate = 268.49] [color={rgb, 255:red, 0; green, 0; blue, 0 }  ][line width=0.75]    (10.93,-3.29) .. controls (6.95,-1.4) and (3.31,-0.3) .. (0,0) .. controls (3.31,0.3) and (6.95,1.4) .. (10.93,3.29)   ;
\draw  [fill={rgb, 255:red, 255; green, 255; blue, 255 }  ,fill opacity=1 ] (84.5,238) -- (221.5,238) -- (221.5,274) -- (84.5,274) -- cycle ;
\draw  [fill={rgb, 255:red, 255; green, 255; blue, 255 }  ,fill opacity=1 ] (242.5,238) -- (379.5,238) -- (379.5,274) -- (242.5,274) -- cycle ;
\draw    (308.5,274.75) -- (308.5,294) ;
\draw    (232.5,293) -- (390.5,293) ;
\draw    (233,293) -- (233.45,310) ;
\draw [shift={(233.5,312)}, rotate = 268.49] [color={rgb, 255:red, 0; green, 0; blue, 0 }  ][line width=0.75]    (10.93,-3.29) .. controls (6.95,-1.4) and (3.31,-0.3) .. (0,0) .. controls (3.31,0.3) and (6.95,1.4) .. (10.93,3.29)   ;
\draw    (390,293) -- (390.45,310) ;
\draw [shift={(390.5,312)}, rotate = 268.49] [color={rgb, 255:red, 0; green, 0; blue, 0 }  ][line width=0.75]    (10.93,-3.29) .. controls (6.95,-1.4) and (3.31,-0.3) .. (0,0) .. controls (3.31,0.3) and (6.95,1.4) .. (10.93,3.29)   ;
\draw  [fill={rgb, 255:red, 255; green, 255; blue, 255 }  ,fill opacity=1 ] (164.5,314) -- (301.5,314) -- (301.5,350) -- (164.5,350) -- cycle ;
\draw  [fill={rgb, 255:red, 255; green, 255; blue, 255 }  ,fill opacity=1 ] (321.5,314) -- (458.5,314) -- (458.5,350) -- (321.5,350) -- cycle ;

\draw (160.75,102.75) node   [align=left] {\begin{minipage}[lt]{59.50000000000001pt}\setlength\topsep{0pt}
{\fontfamily{ptm}\selectfont {\small Abnormal}}
\end{minipage}};
\draw (76.25,180.75) node   [align=left] {\begin{minipage}[lt]{97.58000000000001pt}\setlength\topsep{0pt}
{\fontfamily{ptm}\selectfont {\small Other Lung Diseases}}
\end{minipage}};
\draw (251.75,31.75) node   [align=left] {\begin{minipage}[lt]{97.58000000000001pt}\setlength\topsep{0pt}
{\fontfamily{ptm}\selectfont {\small Scaning Images}}
\end{minipage}};
\draw (328.75,103.75) node   [align=left] {\begin{minipage}[lt]{59.50000000000001pt}\setlength\topsep{0pt}
{\fontfamily{ptm}\selectfont {\small Normal}}
\end{minipage}};
\draw (234.75,178.75) node   [align=left] {\begin{minipage}[lt]{97.58000000000001pt}\setlength\topsep{0pt}
{\fontfamily{ptm}\selectfont  \ \ \ \ \ \ \ \ {\small Pneumonia}}
\end{minipage}};
\draw (159.75,256.75) node   [align=left] {\begin{minipage}[lt]{97.58000000000001pt}\setlength\topsep{0pt}
{\fontfamily{ptm}\selectfont {\small Bacterial Pneumonia}}
\end{minipage}};
\draw (329.75,255.75) node   [align=left] {\begin{minipage}[lt]{97.58000000000001pt}\setlength\topsep{0pt}
{\fontfamily{ptm}\selectfont {\small Viral Pneumonia}}
\end{minipage}};
\draw (236,331.75) node   [align=left] {\begin{minipage}[lt]{64.60000000000001pt}\setlength\topsep{0pt}
{\fontfamily{ptm}\selectfont {\small SARS, MERS}}
\end{minipage}};
\draw (397.25,330.75) node   [align=left] {\begin{minipage}[lt]{56.1pt}\setlength\topsep{0pt}
{\fontfamily{ptm}\selectfont {\small COVID-19}}
\end{minipage}};

\end{tikzpicture}
}
\caption{A semantic representation of the classification of lung diseases}
\label{fig:lung diseases}

\end{figure}
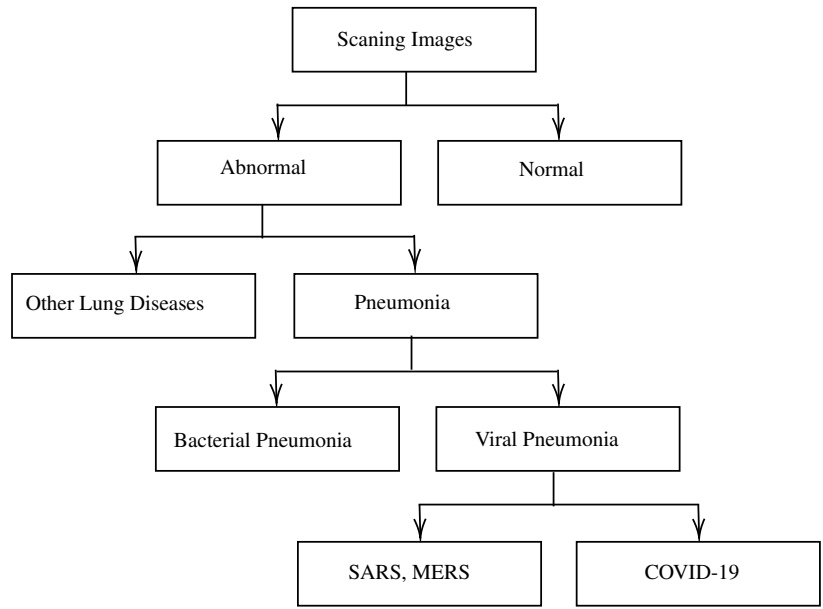

Coronavirus disease 2019 (COVID-19) is caused by the severe acute respiratory syndrome coronavirus 2 (SARS-CoV-2), which first appeared in Wuhan, China in December 2019~\citep{intro} and can easily spread from one person to another person.
The first COVID-19 case reported outside China was on January 13, 2020. A short time later, on January 30, 2020, the World Health Organization (WHO) reported a total of 7,818 confirmed cases worldwide.
The WHO announced the COVID-19 pandemic on March 11, 2020.
Owing to the extremely contagious nature of the virus, the sharp increase in infection and mortality led to the rapid collapse of the world’s socio-economic framework.
As of April 27, 2021, COVID-19 had spread to $223$ countries- there have been $147,539,302$ confirmed cases of COVID-19, including $3,116,444$ deaths~\citep{85}.
These circumstances make it essential to diagnose COVID-19 and isolate the infected people, otherwise the death toll will continue to increase.
Although RT-PCR is widely accepted as a standard diagnostic method, it has some shortcomings.
It requires adequate expertise to collect viral RNA that is extracted from a nasopharyngeal swab of patients.
Also, testing using RT-PCR is time-consuming because it is necessary to maintain stringent laboratory conditions during the testing process.
Therefore, researchers all over the world are seeking a fast, safe and automated method to diagnose COVID-19.
In this regard, researchers have considered two medical imaging techniques, X-ray and CT-scans, to build a fast, accurate, and automated system for COVID-19 detection since both imaging techniques show their supremacy in cases of lung inflammation-related diseases such as COVID-19~\citep{87, 118}.
Ai et al.~\cite{87} manifested that CT scans have a higher sensitivity rate than RT-PCR tests for COVID-19 diagnosis. This research was conducted on $1,014$ patients in Wuhan, China, who experienced both chest CT and RT-PCR tests.
The authors concluded that chest CT tests might be a primary screening tool for detecting COVID-19.
Moreover, it is reported that patients showed abnormalities in chest X-ray images, which is customary for people infected with COVID-19~\citep{123,86}.
A semantic representation of the classification of lung diseases is shown in Figure~\ref{fig:lung diseases}, where we see that an abnormal lung can manifests the COVID-19 infection.

Since last year, there have been many studies investigating machine learning (ML) based architectures in the diagnosis, treatment and follow-up of COVID-19~\cite{126,127}. However, due to the high risk of infection with the virus, medical professionals are particularly at risk.
To control any possible contact with the virus, medical imaging could be the top priority for COVID-19 detection systems.
Although an accurate system would be ideal, diagnosing COVID-19 from medical imaging remains challenging. 
Moreover, a higher detection accuracy and decisive findings are the top requirements for COVID-19 detection.
Since the deep learning (DL)-based architecture for ML in particular is characterized by its impressive recognition performance in medical image classification, DL-based architectures have become attractive candidates for the detection of COVID-19 from chest CT scans and X-rays.
Thus, the architecture based on DL is becoming the key to improving global health risk prevention through reducing epidemiologic risks.
The purpose of this survey is to reveal DL-based architectures proposed by researchers to set up an automated diagnostic system  for COVID-19 using CT scans and X-rays.
The motivation for focusing on the DL-based architecture is that it can introduce key findings related to medical imaging, thereby delivering higher accuracy and key results, which is the goal in detecting COVID-19 from CT scans and X-rays.


In this regard, we have explored several survey papers, including one of the early surveys conducted by Albahri et al.~\cite{124}.
In this survey, the authors looked at automated artificial intelligence (AI) applications based on data mining (DM) and machine learning (ML) algorithms to detect and diagnose Middle East Respiratory Syndrome (MERS)-CoV and Severe Acute Respiratory Syndrome (SARS)-CoV.
What's more, the authors discussed the main features of the coronavirus, the benefits of using machine learning techniques in healthcare, and the limitations of using DM and ML algorithms.
We differ from this survey in many ways, such as the use of incompatible datasets (focusing on MERS-CoV and SAR-CoV instead of COVID-19 CT scans and X-rays), and the use of traditional ML algorithms for classification instead of DL architecture.
Furthermore, in another survey, Albahri et al.~\cite{125} presented a systematic overview of techniques for detecting and classifying COVID-19 from chest CT scans and X-rays for assessment and benchmarking.
The authors conducted a survey of $11$ AI-driven (i.e., traditional ML and DL) research studies to detect and classify COVID-19 using various case studies.
This survey paid limited attention in relation to dataset sources and distribution, dataset preprocessing techniques, direct comparison of various proposed models, and explainability features of the model by activation maps.
However, our survey can also be a more effective alternative as it addresses the shortcomings of the previous survey~\cite{125}.
Next, Shi et al.~\cite{116} addressed AI empowered image acquisition, segmentation, and diagnosis of COVID-19, dealing with X-rays and CT scans.
The difference between this survey and ours is that we not only consider the detection performance of the model but also consider the explainability of the model through visualization.
Considering the explainability feature of the model has always been a key issue in the application of ML in healthcare.
In another survey, Shoeibi et al.~\cite{conp2} discussed a review of the application research of DL technology in COVID-19 diagnosis and automatic lung segmentation, with a focus on work using X-rays and CT scans.
This article also introduces the use of DL architecture to predict the prevalence of coronaviruses around the world.
In addition, Dong et al.~\cite{117} reviewed the imaging characteristics and the use of computational models that have been used for COVID-19 management.
For the detection, treatment, and follow-up of COVID-19, several imaging techniques have been explored, including magnetic resonance imaging (MRI), lung ultrasound, CT, and positron emission tomography-CT (PET/CT).
The quantitative analysis of imaging data using AI is also discussed.
In~\cite{conp2,117}, although some attempts have been made to incorporate image acquisition, image preprocessing, feature extraction and classification paradigms into their survey, a comparison between the performance of the ML-based model and the performance of the radiologist is still lacking.

Our approach is different in many ways to previous survey papers.
In our survey on ML-based COVID-19 detection, we mainly focus on data sources and pre-processing strategies, feature extraction, classification, visual explanation techniques, and detection performance of ML-based models and radiologists.
First, we have explicitly reviewed the dataset sources for CT and X-ray image bench-marking, and then studied the dataset pre-processing techniques.
Then, the different feature extraction techniques that were retained in detecting COVID-19 are discussed.
Next, we have reviewed the classification methods used to detect COVID-19 and continue our exploration by analyzing the visual interpretation techniques implemented in previous methods.
In fact, visually interpretable models make disease detection easier to understand. 
At the end, we also collect information about the radiologist's findings and recommendations for ML-based methods for detecting COVID-19.
None of the review papers~\cite{conp2,116,117,124,125} available tried to examine the relative recognition performance between ML-based methods and human-centered radiologists.
It is necessary to focus on validating the results of the ML-based method with the results of radiologists, as most of the benchmarks were generated from several heterogeneous sources.
To address this previously mentioned deficiency, we report (in Section~\ref{Detection performance between ML-based methods and radiologists}) on the detection performance between ML-based methods and radiologists.
From our survey, we admit that the detection performance of ML-based methods is better than that of radiologists. 
This is the most relevant finding, and perhaps the most important finding in the development of AI-assisted COVID-19 detection from medical imaging.
Moreover, this survey contributes to heath information technology literature, by revealing new insights from using analytical methods. In the survey study, proposed understanding would be offering supportive guidance to healthcare researchers and academics in this sphere in designing and trialing new clinical information systems for detecting and diagnosing any diseases through analyzing relevant CT or X-ray images.
It is anticipated that the overall experiential approach that we introduce in this paper will enable the generation of precise outcomes in detecting diseases and help ensure safe and high-quality clinical operations for all nations. 


\section{Methodology}
\label{sec:methodology}
After selecting the required papers in this survey, we will focus on several important steps followed in different papers for diagnosing COVID-19. 
First, we describe the source and preprocessing techniques of the dataset, as well as their characteristics and attributes. 
Then, a comprehensive review of feature extraction methods and classification techniques is carried out. 
In the end, the results obtained in the existing studied papers along with visualization techniques are discussed.


In this survey, we looked at many research papers from authenticated databases such as the Web of Science, Scopus, Google Scholar, medRxiv, arXiv, and engrXiv. Table~\ref{tab:search} shows the query string/keywords used for searching papers.
We selected paper that used CT scan and/or chest X-ray images, and excluded papers that used lung ultrasound (LU) and/or magnetic resonance imaging (MRI).
We also excluded those papers that did not detect COVID-19 and instead focused on other issues such as the impact of COVID-19 on the cardiovascular system, epidemic prediction, etc.
The papers that we have included mainly used machine learning techniques to detect COVID-19.
After excluding irrelevant papers, we have found $130$ papers in total ($26$ from Web of Science, $29$ from Scopus, $75$ from Google Scholar and different archives). After excluding $32$ duplicates, we finally included $98$ papers in this survey.

\begin{table}[htbp]
    \centering
     \caption{Keywords used in database search}
    \label{tab:search}
    \begin{tabular}{p{3cm}|p{12cm}}
        \hline
        Database & Query string/Keywords  \\
        \hline
        \hline
         Scopus & (TITLE-ABS-KEY ((``COVID-19" OR ``coronavirus" OR ``Corona virus" OR ``coronaviruses" OR ``2019-nCoV" OR ``SARS-CoV" OR ``MERS-CoV" OR ``Severe Acute Respiratory Syndrome" OR ``Middle East Respiratory Syndrome") AND (``X-Ray" OR ``CT-Scan")) AND TITLE-ABS-KEY ((``deep learning" OR ``machine learning" OR ``Artificial Intelligence")))\\
         \hline
         Web of Science & TS=( ( COVID-19 OR coronavirus OR Corona virus OR coronaviruses OR 2019-nCoV OR SARS-CoV OR MERS-CoV OR Severe Acute Respiratory Syndrome OR Middle East Respiratory Syndrome) AND (X-Ray OR CT-Scan) AND (deep learning OR machine learning OR Artificial Intelligence))\\
         \hline
         Google Scholar, medRxiv, arXiv, and engrXiv & COVID-19, X-Ray, CT-Scan, Machine learning, Deep Learning, Artificial Intelligence\\
         \hline
         
         \hline
         \end{tabular}
\end{table}

\section{COVID-19 Imaging Dataset}
\label{sec: COVID-19 Data Set Description}

In this section, we briefly discuss the sources of datasets commonly used in existing works.

\subsection{COVID-19 CT scan dataset}
\label{sec: COVID-19 on CT scan images}

CT scans can produce a precise image of the patient's chest, making them an effective way to observe the condition of the lung.
Table~\ref{tab:ctimage-dataset} lists the publicly available CT scan data sources that the researchers used in their research to find promising COVID-19 detection models.
In addition, some of the CT scan data sources used in previous studies~\citep{4, 21, 42, 49} have not been made public.

\begin{savenotes}
\begin{table*}[htbp]

\resizebox{1\textwidth}{!}{
\begin{threeparttable}[b]
\centering
\caption{Publicly available COVID-19 CT scan datasets}
\label{tab:ctimage-dataset}
\begin{tabular}{p{3.5cm}cp{3cm}p{5.5cm}r}
\hline
Title &	Data source &	Papers & Category & Total Images\\
\hline
\hline
COVID-19 CT segmentation & \romannum{1} & \cite{2}, \cite{65} &COVID-19: 373 & 829 \\
\hline

COVID-19 Database & \romannum{2}	 & \cite{46}, \cite{69} , \cite{85a} &-- & --\\ \hline
     
COVID-CT & \romannum{3} & \cite{60}, \cite{29}, \cite{85a} &COVID-19: 349,\newline Non-COVID: 397 & 746\\ \hline


Lung Segmentation and Candidate Points Generation &	\romannum{4} & \cite{49}, \cite{109a} &COVID-19: 1,252, \newline Non-COVID: 1,230 &2,482\\ \hline

LIDC & \romannum{5} & \cite{12}  &--  &  1,018\\ \hline

CC-19 Dataset & \romannum{6} & \cite{9}  & COVID-19: 28,395 &34,006\\ \hline

Artificial Intelligence in radiology  & \romannum{7} & \cite{85a} & COVID-19: 1,000 &1,000\\ \hline

Chest CT dataset & \romannum{8} & \cite{98a} & COVID-19: 412,\newline Pneumonia: 412,\newline Normal: 412 & 1,240\\ 

\hline
     
\hline
\end{tabular}

\begin{tablenotes}
\item[\romannum{1}] \url{http://medicalsegmentation.com/covid19/}
\item[\romannum{2}] \url{https://www.sirm.org/category/senza-categoria/covid-19/}
\item[\romannum{3}] \url{https://github.com/UCSD-AI4H/COVID-CT}
\item[\romannum{4}] \url{https://www.kaggle.com/arturscussel/lung-segmentation-and-candidate-points-generation}
\item[\romannum{5}] \url{https://doi.org/10.1118/1.3528204}
\item[\romannum{6}] \url{https://github.com/abdkhanstd/COVID-19}
\item[\romannum{7}] \url{https://mosmed.ai/en/}
\item[\romannum{8}] \url{https://data.mendeley.com/datasets/3y55vgckg6/1}
\end{tablenotes}

\end{threeparttable}
}
\end{table*}
\end{savenotes}

\subsection{COVID-19 chest X-ray dataset}
\label{sec: COVID-19 on chest X-rays}

For an automatic disease diagnosis system, medical practitioners have successfully used chest X-rays to detect the innate symptoms of COVID-19 infection, which is the opaque pattern of the lungs~\citep{bbcreports}.
Details of publicly available chest X-ray data sources for COVID-19 detection are shown in Table ~\ref{tab:xray-dataset}.
Also, several articles~\citep{41, 47, 49, 51, 52} used X-ray images data sources that are not publicly available.

\begin{savenotes}
\begin{table*}[htbp]
\resizebox{1\textwidth}{!}{
\begin{threeparttable}[b]
\centering
\caption{Publicly available COVID-19 X-ray image datasets}
\label{tab:xray-dataset}
\begin{tabular}{p{5cm}p{.8cm}p{10cm}p{4.2cm}r}
\hline
Title &	Data source &	Papers & Catagory &Total\\
\hline
\hline
Covid-chestxray-dataset & \romannum{1}	&
\cite{1}, \cite{3}, \cite{7}, \cite{10}, \cite{15}, \cite{18}, \cite{20}, \cite{22}, \cite{23}, \cite{25}, \cite{31}, \cite{32}, \cite{40}, \cite{48}, \cite{49}, \cite{51}, \cite{54}, \cite{57}, \cite{61}, \cite{66}, \cite{67}, \cite{70}, \cite{71}, \cite{72}, \cite{73}, \cite{82}, \cite{83}, \cite{86a}, \cite{87a}, \cite{88a}, \cite{91a}, \cite{95a}, \cite{100a}, \cite{106a}, \cite{110}, \cite{111}, \cite{112}& COVID-19: 132, \newline non-COVID-19: 41 & 173\\
\hline


 
COVID-19 database & \romannum{2} & \cite{1}, \cite{24}, \cite{43}, \cite{51}, \cite{70}, \cite{84}, \cite{86a},  \cite{99a}  & -- &--\\ \hline


COVID-19 Radiography Database & \romannum{3} & \cite{6}, \cite{24}, \cite{25}, \cite{36}, \cite{83}, \cite{86a}, \cite{111} &COVID-19: 219,\newline Normal: 1,341,\newline Pneuomonia: 1,345 & 2,905\\ \hline




COVIDx Dataset & \romannum{4} & \cite{36}  &COVID-19: 589,\newline Normal: 8,851,\newline Pneumonia: 6,053 &15,493\\ \hline

COVID-19 Chest X-ray Dataset Initiative & \romannum{5} & \cite{36}& --  &48\\ \hline


ActualMed COVID-19 Chest X-ray Dataset Initiative & \romannum{6} & \cite{36}  &-- &238\\ \hline


Pneumonia Classification & \romannum{7} & \cite{33} &COVID-19: 90,\newline Pneumonia: 54,\newline Normal: 1,000 &1,144\\ \hline

COVID-19 & \romannum{8} & \cite{16} &COVID-19: 125,\newline  No finding: 500,\newline Pneumonia: 500 & 1,125\\ \hline

COVIDGR-1.0 & \romannum{9} & \cite{13} & COVID-19: 426,\newline Normal: 426 &852\\ \hline

Tubercolosis & \romannum{10} & \cite{7} &COVID-19: 435,\newline Normal: 439,\newline Pneumonia-bacterial: 439,\newline Pneumonia-viral: 439,\newline Tuberculosis: 434 &2,186\\ \hline


Threadreader & \romannum{11} & \cite{86a} &COVID-19: 50 &50\\ \hline


COVID-19-and-pneumonia-scans & \romannum{12} & \cite{89a} &COVID-19: 199,\newline  Normal: 1,965,\newline Viral Pneumonia: 3,723 &5,887\\ \hline

COVID-19-X-rays & \romannum{13} & \cite{91a} &-- &95\\ \hline






\hline
     
\hline
\end{tabular}

\begin{tablenotes}
\item[\romannum{1}] \url{https://github.com/ieee8023/covid-chestxray-dataset}
\item[\romannum{2}] \url{https://www.sirm.org/category/senza-categoria/covid-19/}
\item[\romannum{3}] \url{https://www.kaggle.com/tawsifurrahman/covid19-radiography-database}
\item[\romannum{4}]\url{https://github.com/lindawangg/COVID-Net/blob/master/docs/COVIDx.md?}
\item[\romannum{5}]\url{https://github.com/agchung/Figure1-COVID-chestxray-dataset}
\item[\romannum{6}] \url{https://github.com/agchung/Actualmed-COVID-chestxray-dataset}
\item[\romannum{7}] \url{https://drive.google.com/open?id=1J9I-pPtPfLRGHJ42or4pKO2QASHzLkkj}
\item[\romannum{8}] \url{https://github.com/muhammedtalo/COVID-19}
\item[\romannum{9}] \url{https://github.com/ari-dasci/covidgr}
\item[\romannum{10}] \url{https://drive.google.com/drive/folders/1toMymyHTy0DR_fyE7hjO3LSBGWtVoPNf?usp=sharing}
 \item[\romannum{11}]\url{https://threadreaderapp.com/thread/1243928581983670272.html}
\item[\romannum{12}]\url{https://public.roboflow.ai/classification/covid-19-and-pneumonia-scans}
\item[\romannum{13}]\url{https://www.kaggle.com/andrewmvd/convid19-X-rays}
\end{tablenotes}

\end{threeparttable}
}
\end{table*}
\end{savenotes}

\section{Dataset Preprocessing Methods}
\label{sec: Dataset Preprocessing Methods}

\begin{figure*}[htbp]
\centering
\includegraphics[width=.7\textwidth]{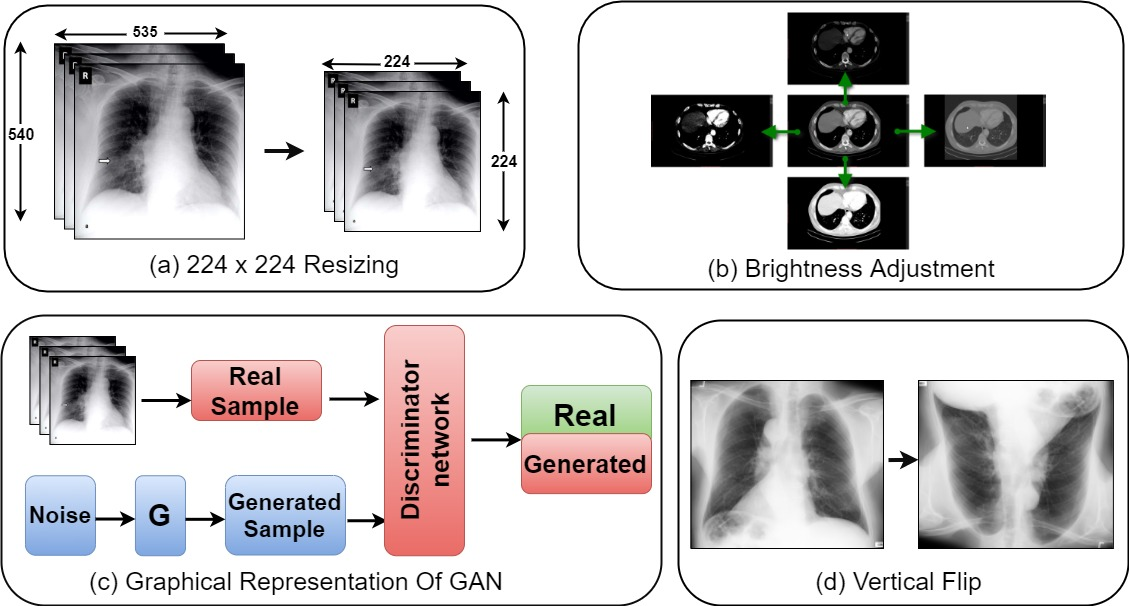}
\caption{Different types of pre-processing techniques}
\label{fig:prepro-methods}
\end{figure*}

Data preprocessing is seen as a strategy for converting raw data into prepared data.
At the same time, data preprocessing is considered an essential diagnostic tool for data because it can fill in missing values, eliminate noisy data, indicate outliers, and find inconsistencies.
Also, the data preprocessing step can make the dataset more versatile, thereby generating a robust DL-based model, thus enhancing the performance of the model.
In this section, we will discuss several data preprocessing methods, which are considered the first step in building a COVID-19 detection model.
Figure~\ref{fig:prepro-methods} shows different types of preprocessing techniques, such as resizing, brightness adjustment, generative adversarial network (GAN), and vertical flipping.

\begin{table}[htbp]
    \centering
      \caption{Different types of preprocessing methods used by the papers}
    \begin{tabular}{p{4cm}p{10cm}}
    \hline
    Preprocessing\newline Methods & Papers\\
    \hline
    \hline
    Resize &\cite{1}, \cite{2}, \cite{6}, \cite{10}, \cite{16}, \cite{18},  \cite{20}, \cite{22}, \cite{23}, \cite{24}, \cite{26}, \cite{28}, \cite{30}, \cite{32}, \cite{43}, \cite{45}, \cite{46}, \cite{48}, \cite{53}, \cite{54}, \cite{56}, \cite{60}, \cite{61}, \cite{67}, \cite{83}, \cite{84}, \cite{85a}, \cite{86a}, \cite{88a}, \cite{95a}, \cite{97a}, \cite{98a}, \cite{100a}, \cite{101a}, \cite{106a}, \cite{108a}, \cite{112}, \cite{114},\cite{118}\\
    \hline
    Flipping or Rotating&\cite{1}, \cite{2}, \cite{15}, \cite{16},    \cite{21}, \cite{22}, \cite{24}, \cite{25}, \cite{27}, \cite{35}, \cite{36}, \cite{38}, \cite{42}, \cite{43}, \cite{46}, \cite{47}, \cite{48}, \cite{51}, \cite{53}, \cite{55}, \cite{57}, \cite{61}, \cite{62}, \cite{65}, \cite{71}, \cite{74}, \cite{82}, \cite{84}, \cite{86a}, \cite{95a}, \cite{98a}, \cite{100a}, \cite{101a}, \cite{102a}, \cite{103a}, \cite{104a}, \cite{107a} , \cite{108a}, \cite{109a}, \cite{112} \cite{114} \\
    \hline
    Scaling or Cropping & \cite{2}, \cite{9}, \cite{18}, \cite{21}, \cite{22}, \cite{25}, \cite{36}, \cite{46}, \cite{48}, \cite{53}, \cite{55}, \cite{56}, \cite{57}, \cite{60}, \cite{61}, \cite{65}, \cite{71}, \cite{74}, \cite{83}, \cite{85a}, \cite{91a}, \cite{95a}, \cite{97a}, \cite{98a}, \cite{102a}, \cite{104a}, \cite{106a}, \cite{112}, \cite{118} \\
    \hline
    Contrast Adjustment & \cite{6}, \cite{21}, \cite{24}, \cite{28}, \cite{30}, \cite{35}, \cite{38}, \cite{41}, \cite{43}, \cite{48}, \cite{52}, \cite{53}, \cite{59}, \cite{68}, \cite{102a}, \cite{103a}  \\
    \hline
    Brightness or\newline Intensity Adjustment & \cite{21}, \cite{36}, \cite{46},      \cite{48}, \cite{52}, \cite{71}, \cite{74}, \cite{82}, \cite{102a}, \cite{108a} \\
    \hline
    GAN & \cite{13}, \cite{44}, \cite{60}, \cite{103a}, \cite{104a}, \cite{114} \\
    
    \hline
    
    \hline
    \end{tabular}
    \label{preprocessing methods used by the papers}
\end{table}

\begin{figure}[htbp]
\centering
\begin{tikzpicture}
  \pie[polar]{3/GAN, 7/Brightness or Intensity Adjustment, 13/Contrast Adjustment, 19/Scaling or Cropping, 28/Flipping or Rotation, 30/Resize }
\end{tikzpicture}
\caption{Ratio of different pre-processing methods}
\label{fig:ratios}
\end{figure}
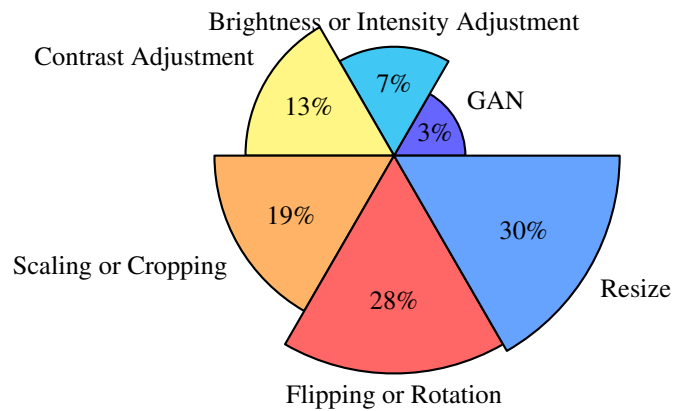

Table~\ref{preprocessing methods used by the papers} summarizes the different preprocessing methods used in previous works.
Resizing is one of the common data preprocessing methods used in $33$ previous works, while flipping was used in $30$ previous works.
Also, Scaling or Cropping, Contrast Adjustment, Brightness or Intensity Adjustment, and GAN were used in a total of $21$, $14$, $8$ and $4$ previous works, respectively. 
Moreover, we have observed that several preprocessing methods were applied in a single model.
For example, some authors used three methods (namely Resizing, Flipping or Rotating, and Scaling or Cropping), while some authors used four methods Flipping or Rotating, Scaling or Cropping, Contrast Adjustment, and Brightness or Intensity Adjustment.
In contrast, some authors have only studied one preprocessing method.
For example, in the paper~\citep{9}, the author used a scale operation. 
In~\cite{16}, resize and rotation operations are considered.
Besides, some researchers used adaptive winner filters to reduce noise ~\citep{59} and affine transformation ~\citep{31}.
The ratio of each pre-processing technique used in the previous study is shown in Figure ~\ref{fig:ratios}, where resizing reaches the highest percentage, while GAN gets the smallest percentage.

\section{Feature Extraction Methods}
\label{sec:Image Feature Extraction}

\begin{figure}[htbp]
    \centering
    \includegraphics[width=.7\textwidth]{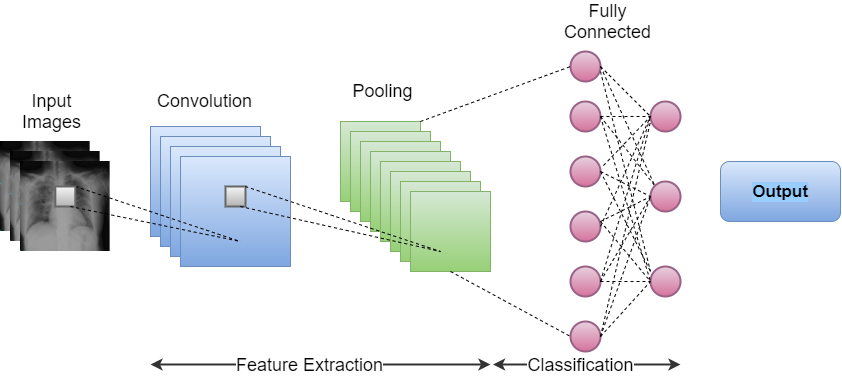}
    \caption{Basic deep learning architecture}
    \label{fig:cnn-arch}
\end{figure}

Feature extraction is the core step of any DL-based image classification model.
Moreover, effective feature extraction is part of the dimensionality reduction process because it can reduce the initial data without missing significant information.
Eventually, the reduced data can be used to develop the model with less machine effort and enhance the speed of learning and generalization steps in the DL-based models.
DL-based classification models comprise two basic steps, feature extraction and classification, in which convolution and pooling operations are combined in the feature extraction process, and fully connected layers become part of the classification.
Figure~\ref{fig:cnn-arch} shows a graphical representation of the DL-based image classification model.

\subsection{Feature extraction methods for CT scan images}
\label{sec: Feature Extraction Methods for CT Image}

To identify patterns from CT scan images between COVID-19 patients and non-COVID-19 patients, the researchers used several related models (i.e., ResNet~\citep{88}, VGG-16~\citep{91}, DenseNet~\citep{90}, etc.).
In Table\ref{Feature Extraction CT}, some of the most used pre-training models are outlined, and among them, ResNet is used extensively, followed by VGG, DenseNet, etc.
In one study~\citep{42}, the authors used ResNet as a pre-trained model to perform binary classification on CT scan images with a depth of $71$ layers and an input image size of $224 \times 224 \times 3$.
In another work~\citep{29}, the authors used DenseNet to improve the computational efficiency by reducing the image size, thereby obtaining an accuracy of $90.61\%$.
As a feature extraction method, the authors~\citep{9} adopted an improved version of Inception v3~\citep{89} named IV3*, and then used the layers of the capsule network to train the extracted features. 
Contrasted with other pre-trained models, the network achieved the highest sensitivity and lowest specificity.
Also, eight different deep learning models used in the paper ~\citep{4} found that the performance of NasNet~\citep{94} and MobileNet~\citep{93} was better than the other six models.

\begin{table}[htbp]
    \centering
    \caption{Feature extraction methods for CT images used by the papers}
    \label{Feature Extraction CT}
    \begin{tabular}{lp{12cm}}

    \hline
    CNN & Papers\\
    \hline
    \hline
    ResNet & \cite{4}, \cite{9}, \cite{12}, \cite{21}, \cite{29}, \cite{42}, \cite{45}, \cite{46}, \cite{55}, \cite{58}, \cite{85a}, \cite{93a}, \cite{97a}, \cite{109a}\\ \hline
    
    VGG & \cite{4}, \cite{9}, \cite{11}, \cite{46}, \cite{53},\cite{55}, \cite{58}, \cite{97a}, \cite{109a} \\
    \hline
    
    DenseNet & \cite{4}, \cite{9}, \cite{21}, \cite{29}, \cite{58}, \cite{109a}\\ \hline
    
    MobileNet & \cite{4}, \cite{9}, \cite{29}, \cite{97a} \\ \hline
    
    SqueezeNet & \cite{42}, \cite{69}, \cite{97a} \\ \hline
    
    Inception & \cite{29}, \cite{68} \\ \hline
    
    CrNet &  \cite{58} \\ \hline
    
    EfficientNet &  \cite{58}, \cite{98a}, \cite{102a} \\ \hline
    
    GoogLeNet &  \cite{46}, \cite{97a} \\ \hline
    
    InceptionResNet &  \cite{4}, \cite{109a} \\ \hline
    
    NasNetMobile  &  \cite{4} \\ \hline
    
    Alex-Net&  \cite{9}, \cite{97a} \\ \hline
    Xception & \cite{97a} \\ 
    
    \hline
    
    \hline
    \end{tabular}
\end{table}

Also, a previous study~\citep{12} used t-SNE for feature visualization and the Least Absolute Shrinkage and Selection Algorithm (LASSO) to identify the 12 most obvious features that distinguish COVID-19 from other pneumonia.
Correspondingly, three additional features were extracted for the region of interest: distance feature, 2D boundary fractal dimension, and 3D gray-scale grid fractal dimension.
As well, in ~\cite{49}, Inception Recurrent Residual Convolutional Neural Network(IRRCNN) was used for feature extraction.

\subsection{Feature extraction methods for chest X-rays}
\label{sec:Feature extraction methods for chest X-rays}

In this section, we have reviewed some state-of-the-art CNN architectures that are used for feature extraction from X-ray images.
From Table~\ref{Feature Extraction methods X-ray images}, the most commonly used CNN architecture is ResNet, followed by DenseNet, VGG, Inception, Xception, etc.
In ~\cite{7}, the author applied 14 CNN architectures to extract features and showed that ResNet has two benefits, namely, the effectiveness of obtaining the highest accuracy and the efficiency of reducing the time of feature extraction.
In another work~\citep{10}, three CNN architectures Xception, ResNet and VGG-16 were explored, but obtained an accuracy of $97\%$ through VGG-16.
Also, in ~\cite{4}, the authors examined eight different CNN architectures using CT scans and X-rays, and NasNet and MobileNet performed better.

Next, we will discuss some custom CNNs for detecting COVID-19.
Some custom CNNs are compatible with visual interpretive methods and perform classification on multi-class or binary class.
An overview of each custom CNN is shown in Table~\ref{tab:custom CNN}.

\begin{table}[htbp!]
    \centering
    \caption{ Feature extraction methods for X-ray images used by the papers}
    \label{Feature Extraction methods X-ray images}
    \begin{tabular}{lp{12.5cm}}
    \hline
    CNN & Papers\\
    \hline
    \hline
    ResNet &  \cite{4}, \cite{7}, \cite{10}, \cite{13}, \cite{14}, \cite{18}, \cite{20}, \cite{22}, \cite{27}, \cite{28}, \cite{31}, \cite{32}, \cite{38}, \cite{40}, \cite{41},\cite{44}, \cite{47}, \cite{51}, \cite{61}, \cite{62}, \cite{66}, \cite{67}, \cite{71}, \cite{73}, \cite{74},\cite{82},\cite{84}, \cite{86a}, \cite{87a}, \cite{91a}, \cite{95a}, \cite{103a}, \cite{106a}, \cite{107a}, \cite{108a}, \cite{112} \\
    \hline
    DenseNet &  \cite{4},  \cite{7}, \cite{14}, \cite{18}, \cite{22}, \cite{23}, \cite{28}, \cite{30}, \cite{31}, \cite{41}, \cite{47}, \cite{48}, \cite{52}, \cite{56}, \cite{66}, \cite{67}, \cite{72}, \cite{74}, \cite{86a},  \cite{95a}, \cite{104a}, \cite{107a}, \cite{108a}, \cite{112}\\
    \hline
    VGG &  \cite{4}, \cite{7}, \cite{10}, \cite{16},  \cite{27}, \cite{28}, \cite{30}, \cite{40}, \cite{47}, \cite{51}, \cite{56}, \cite{66}, \cite{67}, \cite{71}, \cite{83}, \cite{86a}, \cite{87a}, \cite{89a}, \cite{91a}, \cite{95a}, \cite{101a}, \cite{104a}, \cite{112}, \cite{113},\cite{118} \\
    \hline
    Inception &  \cite{14}, \cite{18}, \cite{24}, \cite{33}, \cite{41}, \cite{49}, \cite{56}, \cite{66}, \cite{67}, \cite{91a}, \cite{95a}, \cite{103a}, \cite{104a}, \cite{106a}, \cite{110}, \cite{114} \\
    \hline
    Xception & \cite{7}, \cite{10}, \cite{15}, \cite{30}, \cite{54}, \cite{56}, \cite{61}, \cite{66}, \cite{67}, \cite{91a}, \cite{104a}, \cite{112} \\
    \hline
   InceptionResNet & \cite{4}, \cite{7}, \cite{18}, \cite{22}, \cite{30}, \cite{56}, \cite{66}, \cite{67}, \cite{86a}, \cite{91a}, \cite{95a} \\
   \hline
   NasNet & \cite{4}, \cite{7}, \cite{22}, \cite{23}, \cite{30}, \cite{67}, \cite{104a} \\
   \hline
   
   AlexNet & \cite{7},\cite{44}, \cite{66}, \cite{27}, \cite{28}, \cite{91a}\\
    \hline
   GoogLeNet & \cite{7}, \cite{27}, \cite{40}, \cite{44}, \cite{66}, \cite{91a}\\
   \hline
   SqueezeNet & \cite{41},\cite{44}, \cite{52}, \cite{62}, \cite{86a}, \cite{107a}, \cite{108a}, \cite{111} \\
   \hline
   ShuffleNet & \cite{7}, \cite{52}, \cite{66} \\
   \hline
    MobileNet & \cite{4}, \cite{7}, \cite{30}, \cite{43}, \cite{66}, \cite{86a},  \cite{91a}, \cite{95a}, \cite{99a}, \cite{111} , \cite{112} \\
    \hline
    EfficientNet & \cite{71} \\
    \hline
    CheXNet & \cite{35}\\
    \hline
    CapsNet & \cite{57},\cite{100a}\\
   
    \hline
    
    \hline
    \end{tabular}
\end{table}

\begin{table}[htbp]
\centering
\caption{Summary of custom CNN architectures for COVID-19 detection}
\label{tab:custom CNN}
\begin{tabular}{llclr}
\hline
Custom CNN  &  Interpretability Method & Dataset Type & {Classi-\newline fication} & Accuracy(\%)\\
\hline
\hline

PDCOVIDNet~\citep{1}  & Grad-CAM \& Grad-CAM++  & X-ray & Multi-Class  & $96.58$\\ \hline

CovMUNET~\citep{3} & -- & X-ray & Multi-Class &  $99.41$\\ \hline


COVID-SDNet~\citep{13} & Grad-CAM &X-ray &Binary & $97.37$\\ \hline

COVID-Net~\citep{36}  & Grad-CAM &X-ray &Multi-Class & $93.30$\\ \hline

CovXNet~\citep{26}  & Grad-CAM &X-ray &Multi-Class & $97.40$\\ \hline


CoroNet~\citep{54}  & -- & X-ray &Binary & $89.60$\\ \hline

COVID-CAPS~\citep{57}  & -- & X-ray &Binary & $95.90$\\ \hline

DECAPS~\citep{60} & -- & CT scan & Binary & $87.60$\\  \hline

DarkCovidNet~\citep{87a} & Grad-CAM & X-ray &Multi-Class &$98.08$\\ \hline


Convolutional capsnet~\citep{100a} & -- & X-ray &Binary &$97.24$\\ \hline


\hline

\hline
\end{tabular}
\end{table}

\textbf{PDCOVIDNet} detected COVID-19 from chest X-ray images.
The proposed PDCOVIDNet can change the expansion rate in the parallel stack of the convolutional layer on the CNN, thereby reflecting more distinguishable features, and significantly improving the detection accuracy.
In this article, the authors used $2,905$ chest X-ray images for diagnosis, of which $2,324$ images were used for the training dataset.

\textbf{CovMUNET} is a CNN framework based on multiple losses, which can detect COVID-19 cases from X-ray images.
It contains two branches, the reconstruction branch and classification branch, which calculate two different losses.
In this study, the authors used $5$ fold cross-validation and the dataset size was $6,594$.
Also, the authors who applied CovMUNET for two-class classification (COVID-19 vs non-COVID-19) have achieved significant accuracy, i.e., $99.41\%$, which is better than three-class classification (COVID-19 vs normal vs pneumonia).

\textbf{COVID Smart Data based Network (COVID-SDNet)} is a CNN based classifier that incorporates segmentation, data-augmentation, and data transformations along with a CNN for inference.
The authors introduced us to a high-quality clinical dataset called COVIDGR1.0, which contains $754$ images, of which $377$ are marked as COVID-19.
For the transfer learning method, the researchers adopted ResNet-50 and initialized it with ImageNet weights.

\textbf{COVID-Net} is another CNN architecture that uses a projection-expansion-projection design pattern to detect COVID-19 using chest X-ray images.
The ImageNet dataset is used to obtain the pre-training weights of the proposed COVID-Net, and the COVIDx dataset is applied to the dataset, achieving an accuracy of about $93.3\%$ on the COVIDx dataset.

\textbf{CovXNet} is a multi-dilation convolutional neural network that uses chest X-ray images to automatically detect COVID-19 and other pneumonia, while depthwise convolution is performed on chest X-ray images to extract significant features.
In this work, the authors used a total of $6,161$ images among which the first dataset consists of $5,856$ images ($1,583$ normal, $1,493$ non-COVID viral pneumonia, and $2,780$ bacterial pneumonia) and the second dataset comprises $305$ images, each of the four classes, namely COVID-19, normal, non-COVID viral pneumonia, and bacterial pneumonia.

\textbf{CoroNet} is another CNN-based model that uses the Xception architecture.
In this paper, the authors collected two different publicly available datasets, and then created their datasets to automatically detect COVID-19.
Despite using a small dataset, namely $284$ COVID-19 cases and $30$ normal cases, CoroNet still achieved promising results.
 
\textbf{COVID-CAPS} is a framework based on a capsule network that can identify COVID-19 cases from X-ray images, which consist of four convolutional layers and three capsule layers.
It is worth noting that COVID-CAPS performs better when dealing with smaller datasets.
The number of trainable parameters without pre-training and with pre-training using COVID-CAPS is $295,488$.

\textbf{Detail-Oriented Capsule Networks (DECAPS)} combine Capsule Networks (CapsNets) that are based on ResNet to identify discriminative image features to detect COVID-19 patients using CT images.
A total of $391$ images were marked as COVID-19, while $339$ images were marked as normal.
In this paper, the authors used the DECAPS model with the smallest dataset to obtain an accuracy of $87.60\%$.
Due to the scarcity of sample images, the author applied a conditional adversarial network and other preprocessing techniques, such as Rescaling ($286 \times 286$), and Cropping ($256 \times 256$).
The authors of this article also demonstrated the combination of DECAPS and Peekaboo to improve accuracy.

\textbf{DarkCovidNet} is an architecture that implements $17$ convolutional layers and introduces different filtering on each layer.
It obtained an accuracy of $98.08\%$ and $87.02\%$ for binary and multi-classes.
In addition, the proposed DarkCovidNet provides heat maps that can help radiologists find the affected area on chest X-rays.

\textbf{CapsNet} is built on the capsule network to detect COVID-19 disease by using chest X-ray images.
The proposed method aims to provide a fast and accurate diagnosis for COVID-19, and achieves $97.24\%$ and $84.22\%$ accuracy through binary and multi-class classification, respectively.

\section{Classification Methods}
\label{sec: Classification Methods}

In DL-based architecture, the classification process takes place at the top of a fully connected layer called the softmax layer, and the convolutional layer is performed as a feature extractor in the CNN architecture.
However, some researchers have used pre-trained CNN and support vector machine (SVM) classifiers to achieve improvements~\citep{44, 66}.
Also, in~\citep{14}, the authors incorporate CNN with k-nearest neighbors (k-NN) and a support estimator network, but it requires a lot of data to train.
In the study~\citep{6}, the authors apply the COV-ELM classifier, which applies an extreme learning machine (ELM) to classify COVID-19 from chest X-rays and adjust the network with minimum interference, thereby reducing training time.
An end-to-end web-based detection system with a bagging trees classifier was promoted to simulate the digital clinical pipeline and facilitate the screening of suspicious cases in~\citep{67}.
In~\citep{34}, the authors introduced Adaptive Feature Selection Guided Deep Forest (AFS-DF) as a classification method, which has a higher accuracy than Logistic Regression (LR), Random Forest (RF), Neural Network (NN) and SVM classifiers.
As can be seen from Table~\ref{tab:Classification strategies}, binary classification is most used by researchers, rather than multi-class classification.
However, the binary classification may be ambiguous when detecting COVID-19 because it cannot distinguish between other viral pneumonia and COVID-19.

\begin{table}[htbp]
\centering

\caption{Classification strategies followed by different methods }
\label{tab:Classification strategies}

\begin{tabular}{p{3.5cm}p{11cm}}
\hline
Classification Strategies & Papers \\
\hline
\hline

Binary Class & \cite{4}, \cite{9}, \cite{11}, \cite{13}, \cite{14}, \cite{15}, \cite{18},  \cite{21}, \cite{22}, \cite{23}, \cite{25}, \cite{27}, \cite{28}, \cite{29}, \cite{30}, \cite{32}, \cite{34}, \cite{35}, \cite{42}, \cite{44}, \cite{46}, \cite{48}, \cite{51}, \cite{52}, \cite{53}, \cite{54}, \cite{55}, \cite{57}, \cite{58}, \cite{59}, \cite{60}, \cite{62}, \cite{64}, \cite{65}, \cite{66}, \cite{67}, \cite{68}, \cite{69},  \cite{72}, \cite{83}, \cite{88a}, \cite{97a}, \cite{98a}, \cite{99a}, \cite{100a}, \cite{101a}, \cite{102a}, \cite{105a}, \cite{109a}, \cite{110}, \cite{118}\\
\hline
Multi-Class & \cite{1}, \cite{2}, \cite{3}, \cite{6}, \cite{7}, \cite{10}, \cite{12}, \cite{16}, \cite{20}, \cite{24}, \cite{26}, \cite{31}, \cite{33}, \cite{36}, \cite{38}, \cite{40}, \cite{41}, \cite{43}, \cite{45}, \cite{47}, \cite{49}, \cite{56},  \cite{61}, \cite{66}, \cite{70}, \cite{71}, \cite{73}, \cite{74}, \cite{82}, \cite{84}, \cite{85a}, \cite{86a}, \cite{87a}, \cite{89a}, \cite{91a}, \cite{93a}, \cite{95a}, \cite{103a}, \cite{104a}, \cite{106a}, \cite{107a}, \cite{108a}, \cite{111}, \cite{112}, \cite{113}, \cite{114}\\
\hline

\hline
\end{tabular}

\end{table}


\section{Experimental Results}
\label{Experimental Results}

The evaluation metrics most used in previous studies to assess the performance of DL-based COVID-19 detection systems are accuracy, precision, sensitivity, specificity, and F1-score.
The definitions of accuracy, precision, sensitivity, specificity, and F1-score are as follows:
\[Accuracy=\frac{TP+TN}{TP+FP+FN+TN}\]
\[Precision=\frac{TP}{TP+FP}\]
\[Sensitivity = \frac{TP}{TP+FN}\]  
\[Specificity = \frac{TN}{TN+FP}\]
\[F1 = 2 \times \frac{Precision \times Recall} {Precision + Recall}\]

where $TP$, $TN$, $FP$, and $FN$ stands for true positive, true negative, false positive, and false negative, respectively.
$TP$ refers to the correct classification of the positive class, and the correct classification of the negative class is denoted as $TN$.
$FP$ is a false prediction of positive values, which means that the model classifies an image as COVID-19, but the image does not contain any COVID-19 symptoms.
On the other hand, $FN$ is a false prediction of negative values, for example, the actual COVID-19 image is classified as non-COVID-19.
Among the several methods proposed, some are highly powerful for detecting COVID-19 and perform best in terms of accuracy.
In addition, some proposed methods considered the area under the curve (AUC) to illustrate how precisely the model can predict the results.

\subsection{COVID-19 classification results using CT scans}
\label{sec: Experimental results CT scan images}

Table~\ref{table: Experimental Result using CT images} summarizes the experimental results along with the feature extraction methods using CT images.
We can see that some of the previous methods provided training, testing, and validation ratios, but some methods did not provide a validation ratio. 
In terms of accuracy, the methods using
IRRCNN~\citep{49} and ResNet~\citep{42} obtained the highest scores, $99.56\%$ and $99.4\%$, respectively.
On the other hand, while considering sensitivity as a performance metric, the cGAN~\citep{2} feature extraction method obtained the highest value of $99.97\%$.
The CT scan results used to distinguish COVID-19 have attracted great attention from researchers.
Prior research conducted on CT scans of COVID-19 patients could be characterized using the following features: (1) Detection performance only; (2) Detection performance and region-based learning to label infection and localization of abnormality; (3) Detection performance along with interpretation through visual marker; (4) Verification of results by expert radiologists besides statistical analysis.
Several studies~\citep{21, 29, 34, 46, 59,98a,109a} only provide detection performance through various evaluation metrics such as accuracy, precision, recall, F1-score, and AUC.
On the other hand, some studies~\citep{9, 11, 42, 49, 53,55,64,65} consider the segmentation of the infected lung in addition to the detection performance. In their methods, they tried to localize abnormality in the lungs through region-based learning.
However, to make the model more robust, some studies~\citep{4, 45, 58, 69, 85a} looked at visual interpretation through visual markers.
Due to the lack of sufficient data in the benchmark, certain methods~\citep{12, 60, 68, 93a, 97a, 102a, 105a} verify the statistical results with radiologists. 
These methods will collect data on COVID-19 patients from several hospitals.
Table~\ref{tab:ct survey ratio of experimental evaluation} shows the percentages of different model evaluation criteria, and previous research has focused on this to validate the model.
After reviewing CT scans to examine COVID-19 cases related to DL-based methods, we found that $32\%$ of previous work focused on abnormal lung positioning, while $21\%$ of work focused on radiologist testing and/or verification.
On the other hand, $18\%$ of previous studies covered visual interpretation as well as detection, and $29\%$ of studies performed detection only. 

{\normalsize \tabcolsep=2pt
\begin{longtable}[!htb]{rrcccclp{4cm}}
\caption{Summary of the classification results for CT images}
\label{table: Experimental Result using CT images}\\
\hline
Paper  & Total images & Train,Val,Test ratio (\%) & Train & Val & Test & Result (\%) & Model name\\
\hline
\hline

\endfirsthead
\caption{Summary of the classification results for CT images (continued)}\\
\hline
Paper  & Total images & Train,Val,Test ratio (\%) & Train & Val & Test & Result (\%) & Model name\\
\hline
\hline
\endhead
\hline 
\endfoot
\endlastfoot

\cite{2} & 373 & 80, --, 20 & 300 & -- &73 &99.97 (Se) & cGAN\\ \hline

\cite{4}  & 400 & 80, -- ,20 &	329 & -- &	80 & 95.20 (A)  & NasNetMobile\\ \hline




\cite{9}  & 34,006 & -- & --  & --
& -- & 96.70 (Se)  & Inception \\ \hline

\cite{11}  &4,447 &45,5,50 &2,000 &222 &2,225 & 97 (Se) & VGG\\ \hline

\cite{12}    & 10,250  & --  & -- & -- & -- & 90.19 (Se)  & ResNet\\ \hline

\cite{21}  & 812  & 42, 18, 40  & 341  & 146 
 & 325  & 79.50 (A)  & DenseNet\\ \hline

\cite{29}   & 746  & 52, 14, 25  &425
 &118   &203 & 90.61 (A)  &DenseNet\\ \hline

\cite{34}    & 2,522  & -- & -- & -- & -- & 97.79 (A) & AFS-DF\\ \hline


\cite{42}   &746  &55, 23, 22 &410  & 172
 &164  &99.40 (A)  &ResNet\\ \hline

\cite{45}  & 1,865 & Random Split & 1,725 &270  &320
 &99.40 (AUC) &ResNet\\ \hline

\cite{46}  &6,000 &75,--,25 &4,500 &-- & 1,500 & 98.27 (A)  & Feature Fusion (VGG+ GoogLeNet+ ResNet)\\ \hline

\cite{49}  &420 &45,45,10  &189 &189
&42 &99.56 (A)  &IRRCNN\\ \hline


\cite{53}  & 360 & 80,10,10 &288 &36 &36
& 89.20 (A)  & VGG\\ \hline

\cite{55}  & 3,855 & -- & -- & -- & -- & 95.00 (Se)  & ResNet\\ \hline

\cite{58}  &746 &57,15,28 &425 &119 &208
&83.00 (A) &DenseNet\\ \hline

\cite{59}  &470 &60,--,40 &282 &-- &188
&93.65 (A)  &FFT-Gabor\\ \hline

\cite{60}  &746 &85,--,15 &634 &-- &112
&87.60 (A) &Decapse\\ \hline


\cite{64}  &1,044 &80,10,10 &835 &104 &105
&86.00 (A)  &U-Net\\ \hline

\cite{65} &100 &60,--,40  &60 &-- &40
&83.62 (A)  &MiniSeg\\ \hline

\cite{68}  &453 &--
&-- &-- &-- &89.50 (A)  &Inception\\ \hline

\cite{69}  &783 &56,26,18 & 438 &203 &145
&83.00 (A)  & SqueezeNet\\ \hline

\cite{85a} &2,200 &-- &-- &-- &-- &91.00 (A) &ResNet\\ \hline

\cite{93a} &4,352 &90,--10   &3,916 &-- &436
&98.00 (A) &ResNet\\ \hline

\cite{97a}  &1,020 &80,10,10 &816 &102 &102
&99.4 (A) &ResNet\\ \hline

\cite{98a} &828 &80,10,10 &662 &83 &83
&96.2(AUC) &EfficientNet\\ \hline

\cite{102a}  &132,583 &70,10,20 &92,808 &13,258 &26,517 & 96.00 (A) & EfficientNet \\ \hline

\cite{105a} &538 &-- &-- &-- &-- &97.00 (AUC) &U-Net\\ \hline

\cite{109a} &2,492 &68,17,15 &1,694 &424 &374 &97.00 (AUC) &DenseNet\\ \hline

\hline
\multicolumn{8}{l}{A -- Accuracy, C-V-- Cross-Validation, Se -- Sensitivity}\\
\end{longtable}
}

\begin{table}[!htb]
    \centering
    \caption{Different evaluation criteria used in CT scan based COVID-19 detection research.}
    \label{tab:ct survey ratio of experimental evaluation}
    \begin{tabular}{lp{9cm}l}
    \hline
    Evaluation Criteria & Papers & Total (\%)\\
    \hline
    \hline
    Detection only &  \cite{21}, \cite{29}, \cite{34}, \cite{46}, \cite{54}, \cite{97a} ,\cite{98a}, \cite{109a} 
    & 8 (29\%) \\
    \hline
    Abnormality localization &  \cite{2}, \cite{9},  \cite{11}, \cite{42}, \cite{49}, \cite{53}, \cite{55}, \cite{64}, \cite{65} & 9 (32\%)\\
    \hline
    Visual interpretation &  \cite{4}, \cite{45}, \cite{58}, \cite{69},  \cite{85a} & 5 (18\%)\\
    \hline
    Verify by radiologists &  \cite{12}, \cite{60}, \cite{68}, \cite{93a}, \cite{102a}, \cite{105a} & 6 (21\%) \\
    \hline
    
    \hline
    \end{tabular}
\end{table}

\subsection{COVID-19 classification results using chest X-rays}
\label{Experimental results for chest X-rays}

Table~\ref{tab:experimental result X-ray images} shows an overview of experimental results as well as dataset settings and applied feature extraction methods.
Most previous methods divided the dataset into various ratios of training, testing, and validation, while other methods used cross-validation.
While considering cross-validation, the preceding studies analyzed 10-fold~\citep{6,43,51,67}, 5-fold~\citep{3,7,26,47,70}, and 4-fold~\citep{54} cross-validation.
Compared to 4-fold cross-validation, 10-fold and 5-fold cross-validation provide better results.
Also, accuracy was the most appropriate evaluation index considered by several preceding methods. 
In contrast, few methods consider sensitivity, F1-score, and AUC to demonstrate the potency of these models.
One of the previous studies~\citep{114} used Inception, and its AUC reached $100\%$.
The sensitivity of feature extraction using NASNet-Large~\citep{23} is $100\%$, while the F1-score using the COV-ELM~\citep{6} method is $95\%$.
In addition, in terms of accuracy, NASNetMobile~\citep{4} scored the highest, reaching $100\%$.
Based on the categorization of the experimental evaluation features we discussed in the previous section, we have observed that there are only a few studies~\citep{13,14,62,87a,114} centered on the results of radiologist's verification.
Several studies~\citep{3,16,25,26,41,49,72} claim that their method can detect abnormalities in chest CT scans while improving the accuracy of detection.
Screening with visual interpretation is an important aspect of diagnosing COVID-19. 
Hence, we have seen some studies~\citep{1,4,10,20,22,31,36,47,56,70,71,73,104a} that consider visual markers and test results.
Table~\ref{tab:x-ray survey ratio of experimental evaluation} shows the percentages of different model evaluation criteria performed by previous researchers for model acceptance.
After studying X-ray images to explore COVID-19 facts related to DL-based methods, we observed that $10\%$ of early efforts aimed at the positioning of the infected lung, while $7\%$ of efforts centered on a radiologist’s investigation.
However, $19\%$ of past studies combined visual analysis and detection, while $64\%$ of studies only performed detection.

{\normalsize \tabcolsep=2pt
\begin{longtable}[!htbp]{rrcrrrrl}
\caption{Summary of the classification results for X-rays}
\label{tab:experimental result X-ray images}\\
\hline
Paper  & Total images & Train, Val, Test ratio (\%) & Train & Val & Test & Result(\%) & Model name \\
\hline
\hline

\endfirsthead
\caption{Summary of the classification results for X-rays (continued)}\\
\hline
Paper  & Total images & Train, Val, Test ratio (\%) & Train & Val & Test & Result(\%) & Model name \\
\hline
\hline
\endhead
\hline 
\endfoot
\endlastfoot

\cite{1}  & 2,905 & 80, 10, 10 & 2,324 & 291 & 291 & 96.58 (A) & PDCOVIDNet\\ \hline

\cite{3}  &6,594 &5-fold C-V & -- & -- & -- & 99.41 (A) & CovMUNET\\ \hline

\cite{4}  &400 & 80,--,20  &320 &-- & 80
& 100 (A) & NasNetMobile\\ \hline

\cite{6}  &9,521 &10-fold C-V & -- & -- & -- & 95.00 (F1) &COV-ELM\\ \hline

\cite{7}  &2,186 &5-fold C-V & -- & -- & -- & 98.00 (A) & ResNet\\ \hline

\cite{10}  &375 & 80, 10, 10 &300 &37 &38 & 97.30 (A)  & VGG16\\ \hline
 
 \cite{13}  &754 & 80, 20, -- &603 &150 &-- & 97.37 (A)
&ResNet\\ \hline

\cite{14}  &5,824 &80,--,20 &4,659 &-- &1,165 & 99.49 (A)
&DenseNet\\ \hline

\cite{15}  &1,419 &80,--,20 &1,135 &-- &284 &98.94 (A)
&Xception\\ \hline

\cite{16}   &6,523 &30,40,30 &2,000 &2,523 & 2,000&98.00 (A) &VGG\\ \hline


\cite{18}   &1,302 &60,--,40 &781 &-- &521 &85.10 (Se) &DenseNet\\ \hline

\cite{20}   &15,282 &90,--,10 &13,703 &-- &1,579 &98.06 (A) &ResNet\\ \hline


\cite{22}    &1,214 & 4-fold C-V &--
&-- &-- &98.00 (A) &NASNet-Large\\ \hline

\cite{23}   &3,309 &80,--,20 &2,647 &-- &662 &100 (Se) &NASNet-Large\\ \hline

\cite{24}   &35,500 &88,8,4 &31,340 &2,360
&1,800 &98.00 (A) &Inception\\ \hline

\cite{25}   &1,312 &70,20,10 &918
&131 &263 & 97.40 (A) &CAD system\\ \hline

\cite{26}  &6,161 &5-fold C-V &-- &-- &-- &
97.40 (A) &CovXNet\\ \hline

\cite{27}   &1,864 & -- &-- &-- &-- & 97.54 (A) &ResNet\\ \hline

\cite{28}  &2,271 &70,--,30 & 1,590 &--
&681 &98.90 (A) &DenseNet\\ \hline

\cite{30}  &239 &70,10,20 &167 &22 &20
&98.00 (A) &Residual Att. Net\\ \hline

\cite{31}  &6,297 &27,46,26 &1,591 &2,772 &1,439 &97.10 (A) & DenseNet\\ \hline

\cite{32}   &502 &70,10,20 &399 &3 &100 &88.90 (A) &ResNet\\ \hline

\cite{33}   &1,144 &70,--,30 &801 &-- &343 &89.60 (F1) &Inception\\ \hline

\cite{35}  &6,286 &80,--,20  &5,029 &-- &1,257 &
87.07 (A) &CSEN\\ \hline

\cite{36}   &13,975 &-- &-- &-- &-- &93.30 (A)
&COVID-Net\\ \hline


\cite{38}  &1,764 &70,--,30 & 1,235 &--  &530 &95.12 (A)
&ResNet\\ \hline

\cite{40}   &2,239 &-- &-- &-- &-- &97.01 (A) &DCSL\\ \hline

\cite{41}   &701 &80,--,20 &561 &-- &140 &98.22 (A) &Inception\\ \hline

\cite{43} &3,905 &10-fold C-V &-- &-- & --&99.18(A)
& MobileNet\\ \hline

\cite{44}   &5,824 &80,--,20 &4,659 &-- &1,165 &
99.00 (A) &Resnet\\ \hline

\cite{47}   &16,995 &5-fold C-V &-- &-- &-- &94.80 (A)
&DenseNet\\ \hline

\cite{48}   &414 &70,15,15 &290 &62 &62 &
98.00 (A) &DenseNet\\ \hline

\cite{49}   &5,216 &80,10,10 &4,172 &522 &522 &94.52 (A) & IRRCNN and NABLA-3\\ \hline 

\cite{51}  &455 &10-fold C-V &-- &-- &-- &91.24 (A) & ResNet50+VGG16+CNN\\ \hline

\cite{52}  &537 &70,10,20 &376 &54 &107
&93.50 (A) &MobileNetv2\\ \hline

\cite{54}   &1,300 &4-fold C-V &-- &-- &-- &89.60 (A)
&Xception\\ \hline

\cite{56}  &16,700 &90,--,10 &15,030 &-- &1,670 &99.01 (A) &Inception\\ \hline

\cite{57}  &864 &90,10,-- &777 &87 &-- &95.70 (A)
&COVID-CAPS\\ \hline

\cite{61}  &15,085 &5-fold C-V &-- &-- &-- &
91.40 (A) &Xception+ResNet50V2\\ \hline

\cite{62}  &5,071 &40,--,60 &2,028 &-- &30,426 &97.50 (Se) &ResNet\\ \hline 

\cite{66}   &381 &60,20,20  &228 &76 &76
&95.33 (A) &ResNet\\ \hline

\cite{67}  &274 &10-fold C-V & --&-- &-- &
99.00 (A) &DenseNet\\ \hline

\cite{70}  & 1,277 & 5-fold C-V &-- &-- &-- &
90.13 (A) &VGG\\ \hline

\cite{71}  &13,800 &98,--,2 &13,525 &--
&276 &93.90 (A) &EfficientNet\\ \hline

\cite{72}  &59,937 &80,--,20 &47,950 &-- &11,987&88.04 (AUC) &DenseNet\\ \hline

\cite{73}  &11,663 &91,--,9  &10,613 &-- &1,050 &88.33 (Se) &ResNet\\ \hline

\cite{74}  &15,111 &80,10,10
&12,088 &1,511 &1,511 &89.40 (A) &DenseNet\\ \hline

\cite{82}  &18,567  &90,--,10
&16,714  &-- &1,853 &96.10 (A) &ResNet\\ \hline

\cite{83}  &1,124 &83,--,17 &932 &-- &192 &95.00 (A) &VGG\\ \hline

\cite{84}  &1,122 &92,4,4 &1,052 &35 &35 &72.38 (A) &ResNet\\ \hline

\cite{86a}  &3,487 &72,8,20 &2,510 &279 &698 &72.38 (A) &ResNet\\ \hline

\cite{87a}  &3487 &80,20,-- &900 &225 &-- &98.08 (A) & DarkCovidNet\\ \hline

\cite{88a}  &364 &65,15,20 &233 &56 &75 &96.30 (A) &VGG\\ \hline

\cite{89a}   &396 &80,--,20  &316 &-- &80 &100 (Se) &VGG\\ \hline


\cite{91a} &1,651 &10-fold C-V  &-- &-- &-- &98.75 (A) &VGG\\ \hline



\cite{95a}  &6,086 &80,--,20  &4,869 &-- &1,217 &92.18 (A) &Inception-ResNetV2\\ \hline


\cite{99a}  &3,451 &80,--,20  &2,761 &-- &690 &98.01 (A) &FrMEMs \\ \hline

\cite{100a}  &1,281 &10-fold C-V  &-- &-- &-- &97.24 (A) &Caps-Net \\ \hline

\cite{101a} &284 &70,--,30  &199 &-- &85 &97.97 (A) &VGG \\ \hline

\cite{103a}  &337 &--  & --&-- &-- &89.00 (A) &ResNet \\ \hline

\cite{104a} &15,199 &80,--,20  &12,160 &-- &3,039 &93.00 (A) &VGG \\ \hline

\cite{106a} &6,845 &10-fold C-V &-- &-- &-- &99.96 (A) &Inception \\ \hline

\cite{107a}  &1,725 &80,--,20  &1,390 &-- &335 &98.00 (A) &ResNet \\ \hline

\cite{108a} &5,949 &90,--,10  &5,310 &-- &639 &98.30 (A) &SquzeeNet \\ \hline

\cite{110} &50 &-- & --&-- &-- & 98.99 (A) &Inception \\ \hline

\cite{111} &458 &70,--,30  &321 &-- &137 &99.27 (A) &MobileNetV2 +SqueezeNet\\ \hline

\cite{112} &678 &60,20,20 &406 &136 &136 &98.15 (A) &ResNet \\ \hline

\cite{113} &1,500 &5-fold C-V  &-- &-- &-- &94.93 (A) &VGG \\ \hline

\cite{114}  &572 &5-fold C-V  &-- &-- &-- &100 (AUC) &Inception \\ \hline

\cite{118}   &8,474 &90,--,10 &7,626 &-- &847 &98.60 (A) &VGG\\

\hline

\hline
\multicolumn{8}{l}{A -- Accuracy, C-V-- Cross-Validation, F1 -- F1-score, Se -- Sensitivity}\\

\end{longtable}
}

\begin{table}[htbp!]
    \centering
    \caption{Different evaluation criteria used in X-ray based COVID-19 detection research.}
    \label{tab:x-ray survey ratio of experimental evaluation}
    \resizebox{\textwidth}{!}{
    \begin{tabular}{lp{9cm}l}
    \hline
    Evaluation Criteria & Papers & Total (\%)\\
    \hline
    \hline
    Detection only &  \cite{6}, \cite{7}, \cite{15}, \cite{18}, \cite{23}, \cite{24}, \cite{27}, \cite{28}, \cite{30}, \cite{32}, \cite{33}, \cite{35}, \cite{38}, \cite{40}, \cite{43}, \cite{44}, \cite{48}, \cite{51}, \cite{52}, \cite{54}, \cite{57}, \cite{61}, \cite{66}, \cite{67}, \cite{74}, \cite{82}, \cite{83}, \cite{84}, \cite{86a}, \cite{88a}, \cite{89a}, \cite{91a}, \cite{95a}, \cite{99a}, \cite{100a}, \cite{101a}, \cite{103a}, \cite{106a}, \cite{107a}, \cite{108a}, \cite{110}, \cite{111}, \cite{112}, \cite{113}, \cite{118} 
    & 45 (64\%) \\
    \hline
    Abnormality localization &  \cite{3}, \cite{16},  \cite{25}, \cite{26}, \cite{41}, \cite{49}, \cite{72} & 7 (10\%)\\
    \hline
    Visual interpretation &  \cite{1}, \cite{4}, \cite{10}, \cite{20},  \cite{22},\cite{31},\cite{36},\cite{47},\cite{56},\cite{70},\cite{71},\cite{73},\cite{104a}& 13 (19\%)\\
    \hline
    Verify by radiologists &  \cite{13}, \cite{14}, \cite{62}, \cite{87a}, \cite{114} & 5 (7\%) \\
    \hline
    
    \hline
    \end{tabular}
    }
\end{table}

\subsection{Detection performance between ML-based methods and radiologists}
\label{Detection performance between ML-based methods and radiologists}

Some studies have developed COVID-19 detection models based on ML and verified the results with radiologists (i.e.,~\citep{12, 60, 68, 93a,102a,105a} uses CT images,~\citep{13,14,62,87a,114} uses X-rays).
In the case of CT images, two studies~\citep{93a,105a} involve radiologists reviewing benchmarks and results, but they do not compare the results between ML-based methods and radiologists.
However, other works~\citep{12, 60, 68, 102a} use radiologists to review benchmarks and compare the detection performance between ML-based models and radiologists.
Considering two studies~\citep{12, 60}, the ML-based model gives an AUC score of $0.95$, and when the radiologists are considered, its AUC is $0.85$.
On the other hand, the other two works~\citep{68, 102a} provide extensive evaluations by comparing accuracy, sensitivity, and specificity.
In~\citep{68}, the accuracy, sensitivity, and specificity of the ML-based method are $89.5$, $0.88$, and $0.87$, respectively, and the accuracy, sensitivity, and specificity provided by radiologists are $55.6$, $0.72$, and $0.51$, respectively.
In addition, the ML-based method used in~\citep{102a} has an accuracy of $96\%$, which is $11\%$ higher than the results obtained by radiologists.
This study~\citep{102a} also performs better than radiologists in terms of sensitivity and specificity.
As we have seen from these studies~\citep{12, 60, 68, 102a}, ML-based methods provide better detection performance than independent radiologists.
These results demonstrate the potential superiority of ML-based systems over the detection performance of radiologists.
On the other hand, in the case of X-ray images, some works~\citep{13,14,62,87a,114} create datasets in close collaboration between AI experts and radiologists.
These works did not provide any direct comparison of the detection performance between ML-based methods and radiologists, but radiologists have made some observations that focus on the suitability of ML-based methods and on the performance of the developed system in detecting COVID-19.

\section{Visual Explanation}
\label{ Visual Explanation}

\begin{figure}[htbp]
    \centering
    \includegraphics[width=.6\textwidth]{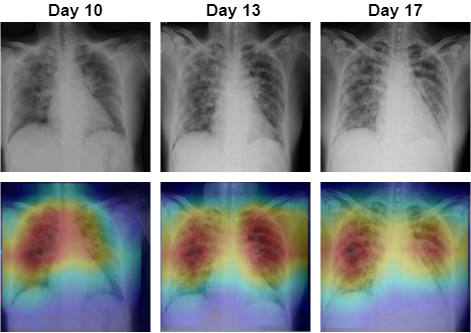}
    \caption{Chest X-ray images of a patent in three points in the upper row and in a lower row their heat maps using Grad-CAM}
    \label{fig:gradcam test}
\end{figure}

\begin{figure*}[!ht]
\centering
\resizebox{8cm}{!}{
\tikzset{every picture/.style={line width=0.75pt}}
\begin{tikzpicture}[x=0.75pt,y=0.75pt,yscale=1,xscale=1]
\begin{axis}[
    ybar stacked,
    bar width=15pt,
    ymin=0,
    nodes near coords,
    legend image code/.code={\draw[#1, draw=none] (0cm,-0.1cm) rectangle (0.2cm,0.2cm);}, 
    legend style={draw=none},
    ylabel={Count},
    symbolic x coords={Grad-CAM, Grad-CAM++, CAM, LIME,LRP},
    xtick=data,
    x tick label style={rotate=45,anchor=east},
    ]
    \addplot+[ybar,  draw=none] plot coordinates {(Grad-CAM, 17) (Grad-CAM++, 3) (CAM, 8) (LIME, 3)(LRP, 1)};
    \addplot+[ybar,  draw=none] plot coordinates {(Grad-CAM, 5) (Grad-CAM++, 0) (CAM, 2) (LIME, 1)(LRP, 0)};
  \legend{\strut X-ray,\strut  CT}
  \end{axis}
\end{tikzpicture}
}

\caption{Visual explanation studies that used machine learning techniques in chest X-rays and CT scans}
\label{Stack bar chart visual explanation}
\end{figure*}
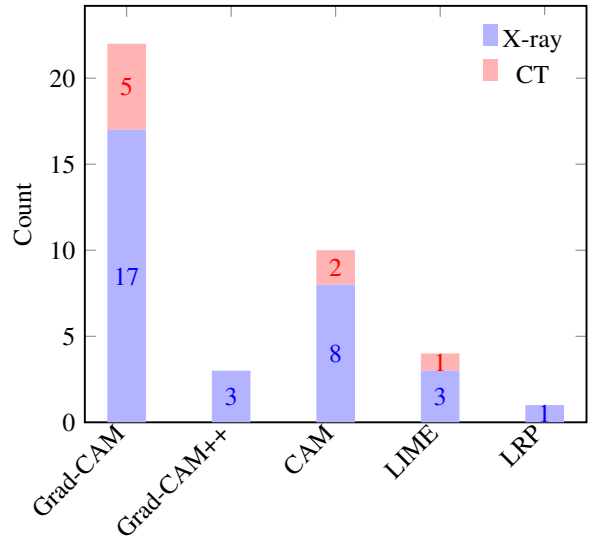

Although the CNN-based architecture can provide excellent results, it is not highly suited to medical diagnostic systems because of the emphasis in such systems on detection performance and visual means to interpret the results.
Some early studies focused on envisioning the behavior of CNN models, and implementing interpretable models through visualization methods that emphasize the importance of identifying classes.
Visualization is an approach for generating class activation maps through heat maps, which can be interpreted as revealing the neural network to make decisions that highlight significant areas in the image.
Many researchers have adopted several visualization methods to interpret the prediction results and identify the key areas of chest X-rays by generating class-discriminating saliency maps.
Some previously used visualization approaches are the Class Activation Map (CAM)~\citep{122}, Gradient-based Class Activation Map (Grad-CAM, Grad-CAM++)~\citep{119}, Layerwise Relevance Propagation (LRP)~\citep{120}, and Local Interpretable Model-Agnostic Explanations (LIME)~\citep{121}, etc.
Fig.~\ref{fig:gradcam test}  shows the output of the heat maps corresponding to the periodic observation of chest X-rays, and how the heat maps intuitively illustrate what the model is concerned about.
As can be seen from Tables~\ref{tab:Interpretability-CT image} and~\ref{tab:Interpretability-X-ray image}, for CT scans and X-rays, Grad-CAM is the most used visualization technique, not CAM.
However, in some studies~\citep{4, 22, 112}, the purpose of exploiting LIME is to obtain visualization, thereby rectifying miss-classification.
Figure~\ref{Stack bar chart visual explanation} summarizes the visual interpretation studies that used machine learning techniques in chest X-rays and CT scans of COVID-19 cases.

\begin{table}[htbp]

\centering
\caption{Interpretability methods used in the CT images based works}
\label{tab:Interpretability-CT image}
\begin{tabular}{ll}
\hline

Interpretability method & Papers\\
\hline
\hline
Grad-CAM & \cite{12}, \cite{45}, \cite{58}, \cite{85a}, \cite{93a} \\ \hline
CAM & \cite{53}, \cite{69}\\ \hline
LIME & \cite{4}\\ 
\hline

\hline
\end{tabular}
\end{table}

\begin{table}[htbp]

\centering
\caption{Interpretability methods used in the X-ray images based works}
\label{tab:Interpretability-X-ray image}
\begin{tabular}{lp{12cm}}
\hline
Interpretability method & Papers\\
\hline
\hline
Grad-CAM & \cite{1}, \cite{6}, \cite{10}, \cite{13}, \cite{15}, \cite{16}, \cite{26}, \cite{31}, \cite{32}, \cite{47}, \cite{56}, \cite{70}, \cite{72}, \cite{73}, \cite{95a}, \cite{104a}, \cite{112} \\ \hline
Grad-CAM++ &  \cite{1}, \cite{20}, \cite{47}\\ \hline
CAM & \cite{10}, \cite{41}, \cite{71}, \cite{72}, \cite{74}, \cite{87a}, \cite{108a}, \cite{114}\\ \hline
LIME & \cite{4}, \cite{22}, \cite{112}\\ \hline
LRP & \cite{47}\\
\hline

\hline
\end{tabular}
\end{table}

\section{Discussion}
\label{Discussion}

In this survey, we studied several diagnostic models to detect COVID-19 and spelled out the characteristics of these diagnostic models.
As mentioned above, Tables~\ref{tab:ctimage-dataset} and~\ref{tab:xray-dataset} show the publicly available chest CT and X-ray datasets, respectively, for COVID-19 detection.
The largest COVID-19 data size of CT images is $28,395$, while for X-ray images it is $589$.
Since images come from multiple sources and different CNN structures have their own specifications, particularly with respect to image size, previous studies have explored data preprocessing methods, as shown in Table~\ref{preprocessing methods used by the papers} and Figure~\ref{fig:ratios}.
The most common preprocessing method used is resizing, and the percentage obtained from GAN is the smallest.
Next, for feature extraction (in Tables~\ref{Feature Extraction CT} and~\ref{Feature Extraction methods X-ray images}), most of the previous research has focused on the use of deep features, and the most widely used CNN architecture for feature extraction is ResNet.
In addition to the commonly used CNN, some previous studies have also developed custom CNN architectures where CovMUNET~\citep{3} shows better performance in terms of accuracy, reaching $99.41\%$.

As can be seen from the experimental results above (in Tables~\ref{table: Experimental Result using CT images} and~\ref{tab:experimental result X-ray images}), it is worth noting that diagnostic models using various methods (i.e. VGG, ResNet, InceptionNetV3, MobileNet v2, Xception, etc.) on CT scans and X-rays give encouraging results for the assessment of COVID-19 patients.
It is evident from this survey that the best accuracy score of detecting COVID-19 based on X-rays is $100\%$~\citep{4}, while the best accuracy found for CT images is $99.56\%$~\citep{49}.
We also observed that the pre-trained CNN model based on transfer learning gained an accuracy of $98.27\%$ for CT images~\citep{46}.
In~\cite{25}, the author proposed a deep learning CAD system to use chest X-ray images to detect COVID-19 and eight other lung diseases, such as atelectasis, infiltration, pneumothorax, mass, effusion, pneumonia, cardiac hypertrophy, and nodules.
A weekly supervised deep learning strategy for detecting and classifying COVID-19 infection from CT images is proposed in~\cite{53}.
The advantage of this model is that it can reduce the need for manually marked CT images.
Although some models reported higher detection accuracy, the datasets used during training were not large enough (i.e.,~\cite{118}).
Finally, some studies have incorporated visualization techniques (i.e., CAM, Grad-CAM, Grad-CAM++, LIME, and LRP) to highlight key regions that are closely related to the predicted outcomes, and the most common visualization technique used is Grad-CAM for CT scan and X-ray models.

Although the application of ML in medical imaging has shown impressive performance in detecting COVID-19, it also has some limitations.
For example, due to the lack of COVID-19 data availability, researchers are processing a limited number of COVID-19 medical images.
Therefore, many public datasets suffer from a class imbalance problem, which causes overfitting, as they have a limited number of COVID-19 cases.
This problem impedes the use of ML to achieve excellent performance because ML techniques do not perform well on skewed datasets.
Another limitation of previous studies is that they did not include clinical symptom information (such as fever, cough, fatigue, etc.) and demographic information (such as age, gender, location, etc.).
Due to the possibility of a mutation in COVID-19, the behaviour of coronavirus is relatively unknown, so using these two types of information can boost the performance of ML-based models.
The third limitation is that most previous studies have paid little attention to the validation of the ground truth.
Therefore, the time between RT-PCR tracking and image information needs to be recorded to establish a benchmark to facilitate effective testing, which helps to make the model more robust.
Another limitation relates to visualization techniques (i.e., CAM, Grad-CAM, etc.) in which activations are highly dispersed and emphasized over irrelevant areas, particularly the mediastinum and shoulder areas in the X-ray image.

To speed up the application of machine learning techniques on a huge number of medical images, a variety of benchmarks of CT scan and X-ray images of COVID-19 have been promulgated throughout the world.
However, one of the challenging tasks is to pre-process the images to make them consistent and aid further analysis, because these images come from various organizations that use heterogeneous scanners.
Another challenge is building a workforce that needs to be built by machine learning experts to develop effective algorithms and by radiologists to thoroughly examine the entire work process.
Although several studies~\citep{12, 60, 68, 93a,102a,105a,13,14,62,87a,114} have developed sophisticated ML-based COVID-19 detection systems and verified the results with radiologists, there is still a lack of the deployment of proposed methods for clinical translation for the diagnosis of COVID-19. 
To maximize the possibility of integrating models into clinical trials to set up cost-effective clinical and technical validation, top-notch benchmarks, external validation and papers with adequate documentation that can be replicated are needed.
As future research, it is suggested that future improvements may combine laboratory test results and image data with clinical findings to better detect and diagnose COVID-19.
We hope that the blend of laboratory test results and clinical findings will contribute to the rapid diagnosis and prognosis of COVID-19.
In addition, it may include the explainability features of the model for fairness and accountability of the model decision.
This can encourage the application of AI-assisted diagnosis of COVID-19 in clinical use.

\section{Conclusion}
\label{Conclusion}

COVID-19 is a highly infectious disease that can quickly affect the lungs.
If we cannot diagnose it quickly and accurately, it may lead to irreparable disasters, including death.
The mortality rate can be reduced by identifying infected patients early and providing them with appropriate treatments. 
This may happen when using a DL-based automatic diagnosis system because it can be diagnosed precisely in a short time.
Also, DL-based diagnostic models that use chest CT scan and X-ray images have a higher potential to reinforce radiologists in rapid COVID-19 detection.
In this survey, we outline the research involving DL-based diagnostic models to detect COVID-19 from CT scans and X-rays, focusing on feature extraction methods, classification, detection performance, and interpretability.
We believe that our survey can provide important insights for medical imaging and help researchers around the world in the fight against the COVID-19 pandemic.
Also, beyond this, developed knowledge presented in this paper could apply to other similar disease diagnosis and detection domains for enhancing quality clinical outcomes.


\section*{COMPLIANCE WITH ETHICS GUIDELINES}
The authors, A. Panday, M. A. Kabir, and N. K. Chowdhury, declare that they have no conflicts of interest.

\noindent
This article does not contain any study materials with human or animal subjects performed by any of the authors.


\begingroup
\let\itshape\upshape
\bibliography{reference}
\endgroup

\end{document}